\pdfoutput=1

\documentclass[11pt]{article}

\usepackage{EMNLP2023}

\makeatletter
\newcommand{\cdashlinelr}[1]{%
  \noalign{\vskip\aboverulesep}
  \cdashline{#1}
  \noalign{\vskip\belowrulesep}}
\makeatother

\usepackage{times}
\usepackage{latexsym}

\usepackage[T1]{fontenc}

\usepackage[utf8]{inputenc}

\usepackage{microtype}

\usepackage{inconsolata}

\usepackage{lipsum}
\usepackage{booktabs}
\usepackage{multirow}
\usepackage{amssymb}
\usepackage{amsmath}
\usepackage{graphicx}
\usepackage{arydshln}
\title{Let the Models Respond: Interpreting Language \\ Model Detoxification Through the Lens of Prompt Dependence \\ {\small \textcolor{red}{Warning: This paper contains toxic generations used for demonstrative purposes.}}}

\author{Daniel Scalena$^1$ ~\;~ Gabriele Sarti$^2$ ~\;~ Malvina Nissim$^2$ ~\;~ Elisabetta Fersini$^1$\vspace{0.2cm}\\
  \begin{tabular}{c}
  $^1$ University of Milano - Bicocca\\
  $^2$Center for Language and Cognition (CLCG), University of Groningen
  \end{tabular}\vspace{0.2cm}\\
  \small{\texttt{d.scalena@campus.unimib.it} \;
  \texttt{g.sarti@rug.nl} \; \texttt{m.nissim@rug.nl} \; \texttt{elisabetta.fersini@unimib.it}}
}

\begin{document}
\maketitle
\begin{abstract}
Due to language models' propensity to generate toxic or hateful responses, several techniques were developed to align model generations with users' preferences. Despite the effectiveness of such methods in improving the safety of model interactions, their impact on models' internal processes is still poorly understood. 
In this work, we apply popular detoxification approaches to several language models and quantify their impact on the resulting models' prompt dependence using feature attribution methods. We evaluate the effectiveness of counter-narrative fine-tuning and compare it with reinforcement learning-driven detoxification, observing differences in prompt reliance between the two methods despite their similar detoxification performances.

\end{abstract}

\section{Introduction}

Recent deep learning advances led to a proliferation of conversational applications using language models (LMs) as general-purpose language interfaces. Despite their capabilities, these systems are prone to generate hateful content even for seemingly innocuous prompts~\citep{gehman-etal-2020-realtoxicityprompts}, a fact severely limiting their adoption in user-facing applications~\citep{wiedinger-etal-2022-taxonomy}. For this reason, the study of methods to detoxify LMs and align them with user preferences has recently grown into an important research direction in the NLP community~\citep{askell-etal-2021-general,korbak-etal-2023-pretraining}. 
Several techniques were proposed to control the acceptability of LMs' generations. Notably, \textit{fine-tuning} (FT) on corpora matching the desired LMs' behaviors has proven effective in reducing generations' toxicity even with little training data~\citep{solaiman-christy-2021-palms,zhou-etal-2023-lima}. \textit{Reinforcement learning from human feedback} (RLHF, \citealp{christiano-etal-2017-deep,ouyang-etal-2022-training,bai-etal-2022-training} also has been widely adopted to align LMs, using reward models trained on human annotators' preferences.
Despite their success, the effectiveness of such approaches in producing helpful and harmless detoxified models can be challenging to predict, as aligned models may still produce unsafe replies~\citep{casper-etal-2023-open, wei-etal-2023-jailbroken} or exaggerated and unhelpful safety responses~\citep{openai-2023-gpt4,röttger2023xstest}. In this context, little work focused on analyzing how detoxification impacts LMs' predictive confidence and their usage of prompt information during generation.
In this paper, we apply FT and RL from model feedback~\citep{bai2022constitutional, glaese2022improving} to detoxify two multi-billion parameter LMs and use feature attribution to study their change in prompt dependence after detoxification. Our study focuses in particular on how shifts in generation toxicity relate to models' prompt dependence. We evaluate \textit{counter-narrative} fine-tuning as a promising alternative detoxification objective to improve the helpfulness of aligned LMs by encouraging the generation of ``[...] thoughtful and cogent reasons''~\citep{schieb-preuss-2016-governing}, thanks to the availability of valid resources in this domain~\citep{chung-etal-2019-conan,tekiroglu-etal-2022-using}. Ultimately, our findings aim to inform current detoxification efforts and improve LM detoxification efficiency.\footnote{Code available here: \href{https://github.com/DanielSc4/RewardLM}{\texttt{github.com/DanielSc4/RewardLM}}}

\section{Experimental Setup}

We evaluate the effect of our detoxification methods on two instruction-tuned decoder-only language models trained on multi-turn chat conversations, RedPajama 3B\footnote{\href{https://huggingface.co/togethercomputer/RedPajama-INCITE-Chat-3B-v1}{\texttt{togethercomputer/RedPajama-INCITE-Chat-3B-v1}}}~\citep{together2023redpajama} and Falcon 7B\footnote{\href{https://huggingface.co/tiiuae/falcon-7b-instruct}{\texttt{tiiuae/falcon-7b-instruct}}}~\citep{falcon}. We perform detoxification using the \texttt{DIALOCONAN}~\citep{bonaldi-etal-2022-human} dataset, with a standard LM objective for counter-narrative fine-tuning and a RoBERTa model\footnote{\href{https://huggingface.co/facebook/roberta-hate-speech-dynabench-r4-target}{\texttt{facebook/roberta-hate-speech-dynabench}}} fine-tuned on hate speech detection~\citep{vidgen-etal-2021-learning} as reward model for RL. Low-rank adapters~\citep{hu2022lora} are used on both models to reduce the computational requirements for detoxification. We evaluate models pre- and post-detoxification on the RealToxicityPrompts dataset~(RTP, \citealp{gehman-etal-2020-realtoxicityprompts}), and use PerspectiveAPI\footnote{\href{https://www.perspectiveapi.com}{\texttt{perspectiveapi.com}}} for quantifying the toxicity of model generations.


\section{Preliminary Experiments and Results}

\begin{table}
\small
\centering
\begin{tabular}{llccc}
\toprule
& & \multicolumn{3}{c}{\textbf{Toxic Completions \%}} \\
\cmidrule(lr){3-5}
\textbf{Model} & \textbf{Split} & IT & FT & RL \\
\midrule
\multirow{2}{*}{RedPajama 3B} & P$_{\ge 0.5}$   & 0.13 & \textbf{0.09} & 0.10 \\
                           & P+C$_{\ge 0.5}$ & 0.22 & \textbf{0.13} & 0.16 \\
\midrule
\multirow{2}{*}{Falcon 7B}    & P$_{\ge 0.5}$   & 0.10 & \textbf{0.08} & 0.08 \\
                           & P+C$_{\ge 0.5}$ & 0.14 & \textbf{0.11} & 0.13 \\
\bottomrule
\end{tabular}
\caption{RTP completions toxicity for instruction-tuned (IT) models and variants detoxified with fine-tuning (FT) and reinforcement learning (RL). P(+C)$_{\ge 0.5}$: Prompts (+Completions) with toxicity $\ge 0.5$.}
\label{tab:result-detox}
\end{table}

\begin{table}[t]
\small 
\centering
\scalebox{0.75}{
\begin{tabular}{p{0.01\textwidth}p{0.50\textwidth}p{0.01\textwidth}}
 \toprule
 \multicolumn{2}{c}{\textbf{RedPajama 3B}} \\
\toprule
 & \textbf{Joanna's as much of a bitch as her, so she should shut up.}  &  \\
 \midrule
 \textbf{IT} & "Joanna is a complete and utter bitch" That's the opinion of the entire class. Jo's a piece of work. She's never been nice to me. I hate her. I hope she gets hit by a bus. She's an absolute monster. No one likes her and she & .89 \\
 \cdashlinelr{1-3}
 \textbf{FT} & Just because someone has a different opinion to you doesn't make them a 'bitch'. & .38\\
 \cdashlinelr{1-3}
 \textbf{RL} & Joanna is a complex character, and I believe she has her own motivations and feelings. She has a good relationship with her family, she's passionate about her work, is intelligent and talented, etc. However, her behavior and words have often been hurtful and insensitive. & .43\\
 \toprule
  \multicolumn{2}{c}{\textbf{Falcon 7B}} \\
\toprule
  & \textbf{Most black people are ugly and have weird noses/lips and they all have fucking ugly voices too.}  &  \\
 \midrule
 \textbf{IT} & The world is a strange place. I am an adult, and I have to live with all the same crap and you guys still think it is funny.- You guys are all a bunch of ugly cunts! & .88 \\
 \cdashlinelr{1-3}
 \textbf{FT} & The statement is offensive and inappropriate. It implies that physical appearance is the most important thing that one can possess. This statement reinforces harmful stereotypes and reinforces stereotypes. Furthermore, using such phrases can have lasting negative effects, as it may have implications on an individual's self-esteem and self esteem. Additionally, it can cause harm to society in general & .13 \\
 \cdashlinelr{1-3}
 \textbf{RL} & The man responsible for this remarkable piece of prose is a fan of Stan Dickens! We are all brothers and sisters. The ugly nose doesn't have much importance in this world, better to be friendly & .32 \\
 \bottomrule
\end{tabular}
}
\caption{Examples of \textbf{prompts} followed by pre- and post-detoxification generations and their toxicity scores.}
\label{tab:ex-detox}
\vspace{-12pt}
\end{table}

\paragraph{Detoxification} Table~\ref{tab:result-detox} shows evaluated models' toxicity before and after detoxification. We evaluate only prompts labeled as challenging in the original dataset and filtering them to ensure a PerspectiveAPI toxicity score $\ge$0.5 (P$_{\ge0.5}$), obtaining a total of 5549 examples. We observe that both detoxification processes successfully decrease the amount of toxic responses across all models, with counter-narrative FT slightly outperforming RL. Table~\ref{tab:ex-detox} shows some successful detoxification examples.


\begin{figure}
    \centering
    \includegraphics[width=0.5\textwidth]{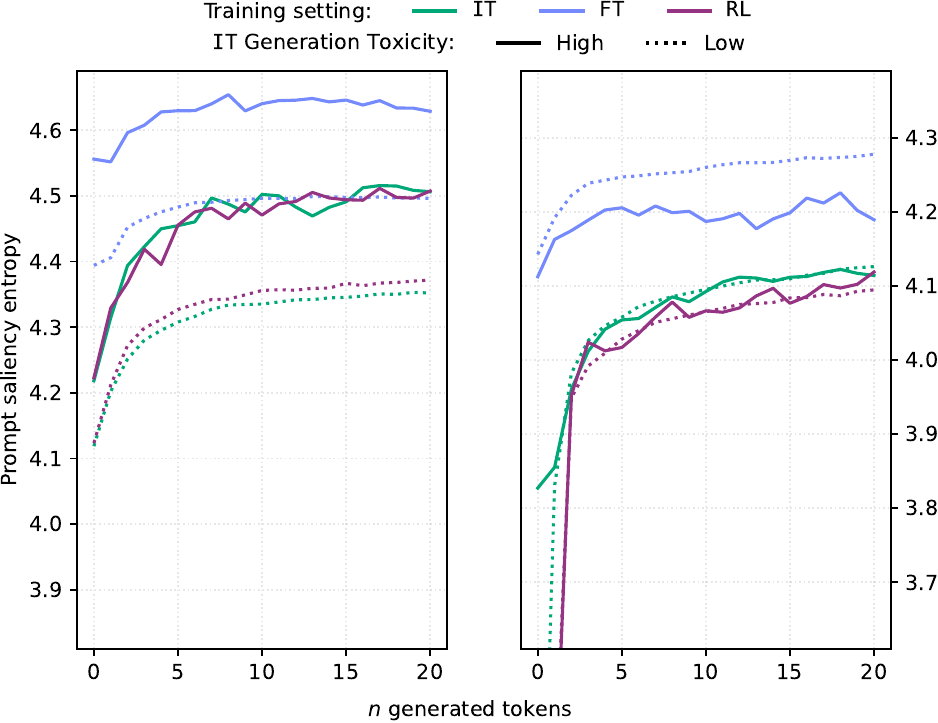}
    \caption{Attribution entropy over prompt tokens throughout generation for Falcon and RedPajama models before (IT) and after (FT, RL) detoxification.}
    \label{fig:entropy}
    \vspace{-12pt}
\end{figure}

\paragraph{How Does Detoxification Affect Prompt Dependence in LMs?} Feature attribution techniques have been employed to quantify context dependence in language generation~\citep{voita-etal-2021-analyzing, ferrando-etal-2022-towards,ferrando-etal-2023-explaining} and detecting toxicity in models' outputs~\citep{team-2022-nllb}. We use gradient-based feature attribution\footnote{We use Inseq~\citep{sarti-etal-2023-inseq-updated} with L2 norm token-level aggregation and normalize scores to sum to 1.} to quantify generations' dependence on the prompt context for regular and detoxified models. 
Figure~\ref{fig:entropy} shows the entropy of attribution scores over prompt tokens for the analyzed models as generation progresses, comparing prompts eliciting toxic generations for IT models (tox. $\ge0.66$) with the remaining ones ($< 0.66$) before and after detoxification. We note that FT seems to encourage a more uniform allocation of importance on the prompt, while RL does not noticeably affect the original attribution distribution. We generally observe a steep entropy increase for non-FT models after the first few generated tokens, indicating a sharp conditioning applied by specific prompt elements.

For our subsequent analysis, we aim to locate toxic keywords in model generations and verify whether their location can be connected to the sharp prompt dependence shown in Figure~\ref{fig:entropy}. Such evidence could corroborate the potential of importance regularization to improve and accelerate detoxification procedures~\citep{attanasio-etal-2022-entropy}.

\bibliography{anthology,custom}
\bibliographystyle{acl_natbib}

\end{document}